\pdfoutput=1

\documentclass[11pt]{article}

\usepackage{acl}

\usepackage{times}
\usepackage{latexsym}

\usepackage[T1]{fontenc}

\usepackage{subfig}
\usepackage{float}
\usepackage{graphicx}
\usepackage{enumitem}

\usepackage[utf8]{inputenc}

\usepackage{microtype}

\definecolor{my-dark-green}{HTML}{38946b}
\definecolor{my-dark-orange}{HTML}{e05d06}
\definecolor{my-dark-purple}{HTML}{4665a0}
\definecolor{my-dark-pink}{HTML}{be2784}
\definecolor{my-dark-yellow}{HTML}{8dc224}

\title{`\emph{Tecnologica cosa}': \\Modeling Storyteller Personalities in Boccaccio's \emph{Decameron}} 

\author{A. Feder Cooper\thanks{$\;$ Corresponding author; \texttt{afc78@cornell.edu}} \and Maria Antoniak \and Christopher De Sa \and David Mimno\\
  Bowers College of Computing and Information Science, Cornell University \\\AND
  Marilyn Migiel \\
  Department of Romance Studies, Cornell University \\}

\begin{document}
\maketitle
\begin{abstract}
We explore Boccaccio's \emph{Decameron} to see how digital humanities tools can be used for tasks that have limited data in a language no longer in contemporary use: medieval Italian. We focus our analysis on the question: Do the different storytellers in the text exhibit distinct personalities? To answer this question, we curate and release a dataset based on the authoritative edition of the text. We use supervised classification methods to predict storytellers based on the stories they tell, confirming the difficulty of the task, and demonstrate that topic modeling can extract thematic storyteller ``profiles." 
\end{abstract}

\section{Introduction}

The \emph{Decameron} is a masterpiece of medieval Italian literature. Completed by 1353, the text is often referred to as ``l'umana commedia'' (The Human Comedy),\footnote{The first words of the Proem are ``Umana cosa,'' which roughly translate to ``It is human''  or ``human quality''~\cite{mcwilliam1975boccaccio}; they immediately underscore the secular focus.} a name meant to strike a contrast in subject matter (and a parity in importance) with Dante's well-known ``divina commedia'' (Divine Comedy)~\cite{branca1975boccaccio}. 
In a structure similar to Geoffrey Chaucer's \emph{The Canterbury Tales}, it is a collection of 100 stories (\emph{novelle}) woven together in the context of a frame tale: an honorable brigade (\emph{brigata}) composed of 7 women and 3 men who have fled the ravages of plague in Florence to the relative seclusion of the Tuscan countryside.\footnote{The description of the chaos inflicted by plague in Florence has led to renewed international interest in the text~\cite{findlen2020covid, marcus2020covid, nyt2020decameron}.} The 100 \emph{novelle} are told by the 10 \emph{brigata} members over 10 days, with each day assigned a theme spanning matters of love, wit, and trickery.

While scholarship for \emph{The Canterbury Tales} has engaged with both the stories and the storytellers~\cite{kittredge1915chaucer,lawton1985chaucer,ginsberg2015tellers}, storyteller identity has received relatively less attention in the \emph{Decameron}. Instead, literary research has tended to address themes and stories~\cite{migiel2004rhetoric,migiel2015ethical,marcus1979form}. Treatment of storyteller identity has thus far been sparse, perhaps due to storyteller personalities seeming generally\footnote{``Generally'' should be taken very generously; we do not intend to eclipse or elide the small yet rich corpus of scholarship that has either directly~\cite{marafioti2001lauretta, grossi1991filomena} or indirectly~\cite{richardson1978ghibelline} discussed storytellers.} difficult to distinguish at a high level via close reading.\footnote{This is debatable, but perhaps true in comparison to \emph{The Canterbury Tales}. For over 100 years scholars have investigated pilgrim personalities. See~\citet{kittredge1915chaucer}, at p. 155, ``The Pilgrims do not exist for the sake of the stories, but \emph{vice versa}. ... [T]he stories are merely long speeches expressing, directly or indirectly, the characters of the several persons.''}

We therefore ask: \textbf{Do the members of the \emph{brigata} exhibit distinct storytelling personalities?} 
We emphasize that this is not a question of authorship, as the text is attributed to Boccaccio alone, but rather one of thematic and stylistic differences among the fictional characters he depicts. To approach this question, we use computational tools to elicit patterns from the text---patterns that may have thus far remained elusive to scholars and could help constitute unique storyteller identities. 

This case study highlights several challenges for digital humanities research.
While the \emph{Decameron} is a popular and well-studied text, it is written in medieval Italian, for which there are few language modeling resources; this forces us to rely on language-agnostic methods like classification and topic modeling (Section \ref{sec:case}). Moreover, while some digitized resources do exist~\cite{decameronweb, decameron2003online}, the text required multiple rounds of curation to be used for a computational study. To facilitate future digital \emph{Decameron} scholarship, we release our user-friendly digital version (Section \ref{sec:data}).\footnote{\url{https://github.com/pasta41/decameron}} In order to build a training corpus for this domain, similar curation will be necessary for other digitized medieval Italian texts, including additional works by Boccaccio and authors such as Dante and Petrarch. 


Taken together, our classification and topic modeling results support existing humanist scholarship concerning storyteller identity and suggest new questions for further inquiry. More broadly, our work here serves as preliminary evidence that such tools can be useful for highly specialized academic digital humanities work---in non-English and non-standard (e.g., bygone language variant) domains.


\section{Related Work}

As one progresses through the 10 days of the \emph{Decameron}, different members of the \emph{brigata} seem to develop distinct storytelling personalities.
Dioneo frequently tells bawdy tales, pushing the bounds of decorum. Emilia seems like she does not quite fit in with the rest of the group, 
and in fact may be (though we cannot be certain) an actual political outsider---the sole Ghibelline in the group of Guelphs~\cite{richardson1978ghibelline}. Lauretta can perhaps be cast as a ``bearer of bad news,'' according to~\citet{marafioti2001lauretta}.
Notwithstanding such stand-alone examples, scholarship has not addressed whether each of the 10 \emph{brigata} members have clearly identifiable storytelling personalities. 
As there is no literary consensus, we apply computational tools to extract patterns that might be difficult for human readers to elicit.

Prior work in digital humanities has studied a variety of narrative questions across corpora of multiple texts. For example, research in cultural analytics has compared narrative structure~\cite{chambers2009unsupervised,Pichotta2016LearningSS,Goyal2010AutomaticallyPP}, character arcs and relationships~\cite{bamman2013learning,iyyer2016feuding}, and authorship attribution~\cite{hoover2004testing}. Authorship can also be modeled as a latent factor in topic models~\cite{rosenzvi2004authortopic}. We do not explicitly model authors (or in our case, narrators) but instead rely on a simpler model to extract cross-cutting, interpretative themes. Moreover, unlike studies of focalization~\cite{genette1983focalization}, we make no attempt to model the perspective or views of a character, but rather simply ask if characters are in any way distinguishable.

In computational studies that similarly focus on sections of a single work,~\citet{wang2019casting} compare sections of Italo Calvino's \textit{Invisible Cities} and~\citet{brooke2015wasteland} investigate distinguishing narrative voices in T.S. Eliot's \textit{The Waste Land}. \citet{wang2019casting} circumvent data size limitations by relying on large, pretrained contextual models to cluster the cities and compare thematic patterns; \citet{brooke2015wasteland} rely on preexisting tools that elicit English-language features, including parts of speech and verb tense. Such pretrained models and featurization tools, while available for modern Italian~\cite{polignano2019alberto}, are unavailable for the quite different medieval Italian of the \emph{Decameron}~\cite{salvi2010antico, dardano2012sintassi}. We instead use language-agnostic computational tools for our experiments, which come with the added benefit of interpretability (Section \ref{sec:case}). 

\section{Curating a \emph{Decameron} Dataset} \label{sec:data}

We constructed a \texttt{json} dataset of the \emph{Decameron} from an XML version hosted online by the Sapienza University of Rome~\cite{decameron2003online}. This digitized version is based on Vittore Branca's authoritative text~\cite{branca2014decameron}, and was published online in 2003 in the (at the time standard) TEI P4 format~\cite{tei}. TEI P4 contains a variety of metadata that interrupt contiguous portions of Boccaccio's text, which does not make the format amenable to commonly-used tools. We therefore spent considerable time simplifying this format to be more easily manipulable for modern data analysis. We manually and repeatedly verified that our curation process retained the integrity of the text.\footnote{We document the process in our repository README.} Where appropriate, we added metadata to annotate \emph{novelle}, such as the \emph{novella} storyteller, which was absent in the existing online version. Unlike the TEI P4 format, we avoid placing these metadata within the text of individual \emph{novelle}, and provide scripts for those that wish to remove these metadata for their analyses. 

We release this dataset publicly. Our hope is that our online version of Branca's text will be more accessible to scholars of medieval Italian interested in engaging with digital tools, as the simplicity of our format should lower the barrier to entry for both computational and humanist scholars interested in the \emph{Decameron}.


\section{Case Study: Constructing Storyteller Profiles in Boccaccio's \emph{Decameron}} \label{sec:case}
We use the problem of \emph{Decameron} storyteller identity as a case study for exploring the challenges and opportunities for using digital humanities techniques in specialized literary domains. In particular, \textbf{we investigate how such tools can be useful for 1) a small corpus containing a single text 2) modeling language that is no longer in contemporary use.} While the question we ask is specific to our chosen domain---the \emph{Decameron} and medieval Italian---we believe that the lessons we can derive are applicable to other scholarly digital humanities tasks with these same defining elements. 

\subsection{Problem Formulation} \label{sec:model}
\citet{wang2019casting} were able to use pretrained contextual models like BERT~\cite{devlin2019bert} for modeling a small, single-text English corpus, analogous tools are not available for studying the \emph{Decameron}. While there is a modern Italian version of BERT (AlBERTO, trained on Twitter data~\cite{polignano2019alberto}), medieval Italian orthography and morphology are sufficiently different to contraindicate its use~\cite{salvi2010antico, dardano2012sintassi}.
Moreover, such pretrained contextual ML models are difficult to interpret, and our goal is to assist humanist scholars in close-reading analysis. Learning about storyteller identity is not just about classification; we already know authoritatively who told which story. Rather, we would like to explain \emph{how} our models distinguish among storytellers. Based on this goal, and the constraints we highlight in Section \ref{sec:case}, we choose two language-agnostic, interpretable approaches: logistic regression to try to classify storytellers based on their \emph{novelle} and topic modeling to model storytellers as distributions of lexical themes.

\begin{table*}[t]
    \scriptsize
    \begin{center}
    \begin{tabular}{p{0.7cm}p{1cm}p{1cm}p{1cm}p{1.1cm}p{1cm}p{1cm}p{1cm}p{1cm}p{1cm}p{1cm}}
        & \textbf{Panfilo} & \textbf{Neifile} & \textbf{Filomena} & \textbf{Dioneo} & \textbf{Fiammetta} & \textbf{Emilia} & \textbf{Filostrato} & \textbf{Lauretta} & \textbf{Elissa} & \textbf{Pampinea} \\
        \hline
        \noalign{\vskip 0.5mm}   
        \textcolor{my-dark-green}{\textbf{Highest}} & gentili & veder & amico & famigliare & meco & basciò & fama & gentile & figliuolo & freddo \\
        & compagni & onesta & parenti & conoscere & cuore & vivo & caldo & messere & cavalli & torre \\
        & bocca & gentil & ciascun & primieramente & amava & accidente & cavallo & talvolta & liberamente & fante \\
        & figliuol & credeva & cautamente & signor & nell'animo & maravigliò & oimè & diè & figliuoli & amante \\
        & vicina & vicini & dico & pose & venendo & buone & malvagia & belle & veggendosi & reina \\
        \hline
        \noalign{\vskip 0.5mm}  
        \textcolor{my-dark-purple}{\textbf{Lowest}} & freddo & peccato & famigliare & valente & cavallo & vedi & re & incontanente & dormire & veramente \\
        & buone & disidero & sentito & morte & tornare & uom & speranza & animo & allato & ricco \\
        & figliuolo & occhi & porta & corte & signor & famiglia & gentili & medesima & mille & tavola \\
        & aperto & cavaliere & tavola & madre & signore & partito & figliuoli & vedendo & ciascun & bisogno \\
        & pianamente & troppo & corte & cavaliere & figliuoli & cara & amor & fante & giovani & morto \\
       \hline
    \end{tabular}
    \end{center}
    \caption{The words with highest and lowest PMI for each storyteller (higher scores indicate stronger associations).}
    \label{table:pmi}
\end{table*}

\subsection{Modeling Storytellers using Logistic Regression} \label{sec:logr}

\begin{figure}[ht!]
    \begin{center}
    \subfloat[
        \label{subfig:f1_10-class}]{%
      \includegraphics[width=3in]{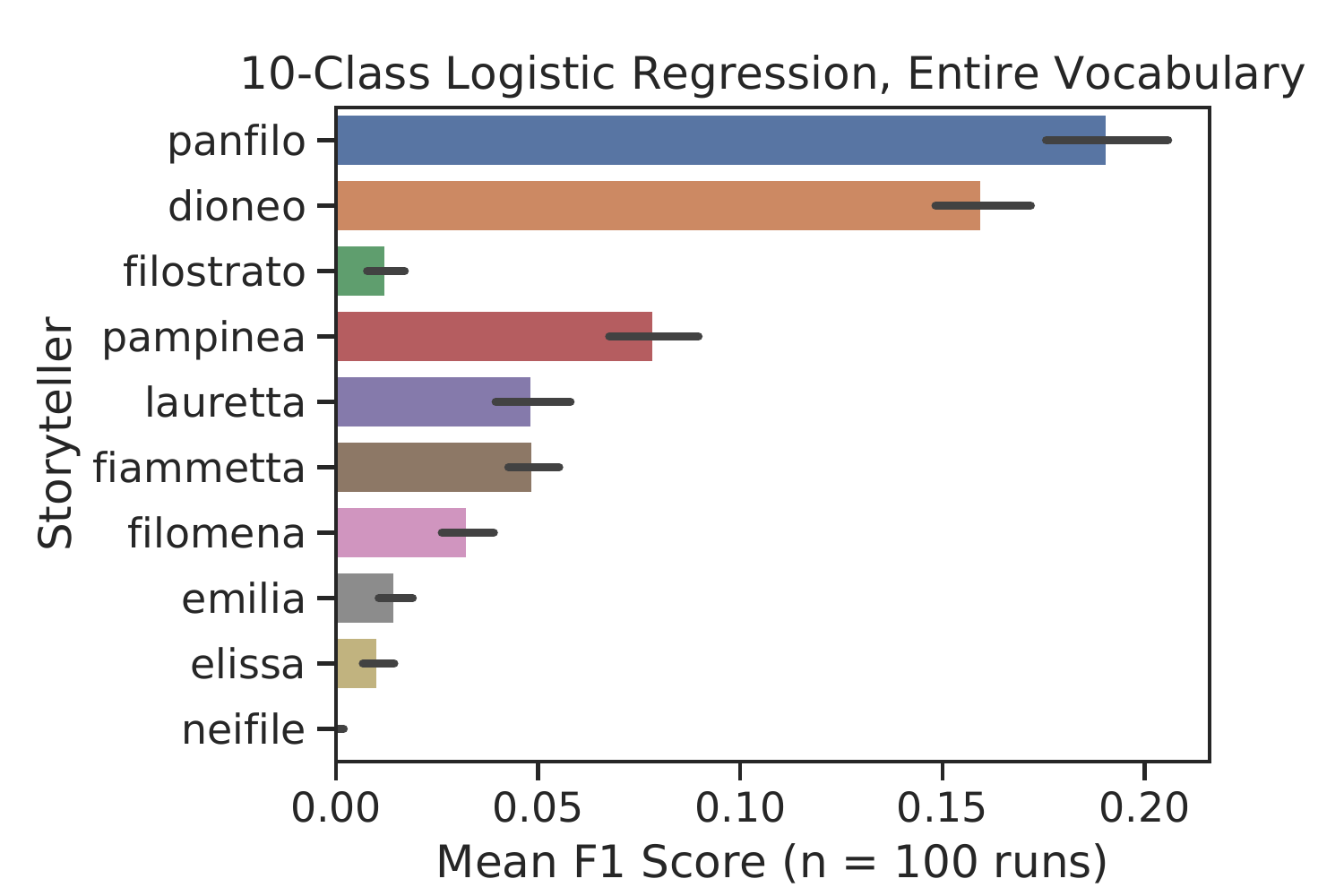}
    }
    \hfill
    \subfloat[
        \label{subfig:f1_10-class-top-100}]{%
      \includegraphics[width=3in]{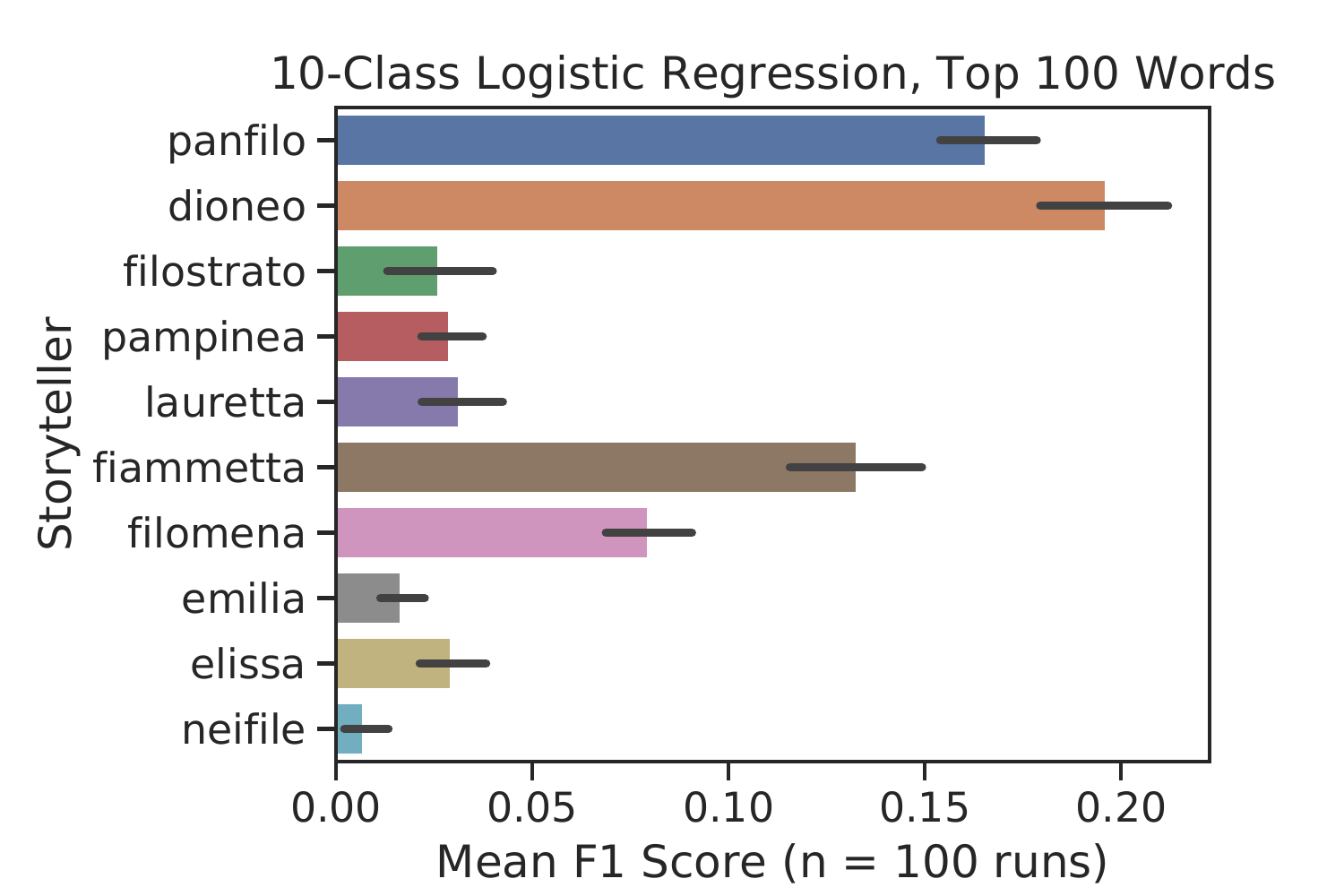}
    }
    \caption{Mean F1 scores for classifying \emph{novelle} by narrator in 10-class logistic regression. For both experiments, $n=100$. Using the \emph{Decameron}'s whole vocabulary (\ref{subfig:f1_10-class}), the model can identify Panfilo and Dioneo better than random. When we restrict the text to only contain instances of the 100 most frequent words (\ref{subfig:f1_10-class-top-100}), the model is additionally able to identify Fiammetta.}
    \label{fig:f1}
    \end{center}
\end{figure}

We first attempt to see if the storytellers can be identified from the \emph{novelle} they tell. 
We train a logistic regression model for this classification task. 
For our training data, we divide each \emph{novella} into 100-word chunks (converted to TF-IDF vectors) with the corresponding storyteller as the label. This results in a 10-class logistic regression problem, using an 80/20 train/test split where we ensure that we have equal representation of each storyteller in both sets (i.e., each storyteller tells 10 \emph{novelle}, comprising the 100 \emph{novelle} total; we randomly sample 8 \emph{novelle} for each storyteller in train; the remaining 2 for each in test). We train our model 100 times, with variation coming from randomly sampling the \emph{novelle}. Results are shown in Figure \ref{subfig:f1_10-class}. 

Since there are 10 storytellers, in order to classify better-than-random, F1 scores would need to be $>0.1$. There are only two storytellers, who are both men, that pass this threshold consistently: Panfilo and Dioneo. It is perhaps unsurprising that this is true for Dioneo; he alone among the \emph{brigata} has the special privilege of deviating from the Day's storytelling theme---a privilege he typically exercises to talk about sex. It is however less clear to us why Panfilo stands out in our results, which suggests a potential direction for future research. 

We then re-ran this experiment, pre-processing the \emph{Decameron} to only contain the 100 most frequently used words in the vocabulary. In addition to Panfilo and Dioneo, this model is also able to identify Fiammetta, one of the seven women, better than random (Figure \ref{subfig:f1_10-class-top-100}). This, too, suggests lines of further investigation, as it is unclear why Fiammetta is more identifiable than the other women.\footnote{We repeated these experiments using one-versus-rest logistic regression to test if each storyteller is distinguishable compared to the other 9. The results were comparable to those presented in Figure \ref{fig:f1}.} 

We probe our classification results by extracting the words with highest and lowest pointwise mutual information (PMI) for each \textit{brigata} member. 
This metric uncovers lexical associations with each narrator in comparison to all the narrators.
Given a word $w$ and a narrator $n$, 
$PMI(w;n) = log \frac{p(w|n)}{p(w)}$.
To improve interpretability, we remove words that occur fewer than five times for each narrator, and we manually remove stopwords.\footnote{[\textit{è, che, la, quale, e, di, fu, le, per, col, aveva, avere, ha, il, lo, gli, i, de, in, ciò, ho}]}
Table \ref{table:pmi} shows that despite our low classifier performance, lexical differences between the storytellers are interpretable. Neifile, whom our classifier completely misses, scores high for words that signify honorability (e.g., \emph{onesta}), while low for words that connote the opposite (e.g., \emph{peccato}). Filostrato, who reveals his personal heartbreak, scores low for words concerning love and hope (e.g., \emph{amor}, \emph{speranza}).\footnote{Notably, PMI scores do not incorporate semantics. As a result, two words that have the same semantic meaning but different morphology can have very different scores. For example, Panfilo's \textit{novelle} have high PMI for \textit{figliuol} but low PMI for \textit{figliuolo}--words that both mean ``son'' (in the medieval variant of the modern ``figliolo'').}  






\subsection{Modeling Storytellers using Topic Distributions} \label{sec:lda}

Since we were not able to generally distinguish storytellers via classification, as a second experiment we use latent Dirichlet allocation (LDA)~\cite{blei2003lda} to model each \emph{novella} as a distribution of topics. We group the results by storyteller to see if the distributions of \emph{novella} topics are distinguishable for each of the 10 members of the \emph{brigata}. In other words, we can view per-storyteller topic distributions as storyteller ``profiles''---patterns that may indicate unique thematic features of particular \emph{brigata} members.

To perform this analysis, we used a Python wrapper for MALLET~\cite{antoniak2021mallet, mccallum2002MALLET}. We used this framework because its implementation of LDA uses Gibbs sampling~\cite{geman1984gibbs}, an exact MCMC sampling method that has popularly been observed to have better performance for small datasets than inexact, variational inference-based implementations. 
We train our model with $k=20$ topics\footnote{We tried different $k$ and found that $20$ resulted in the most interpretable, overarching topics for our small dataset.} and allow hyperparameter optimization. Before training, we lowercase the text and process each \emph{novella} to create documents of 200 words each; we remove a custom list of common Italian stop words, and if the resulting document is fewer than 20 words long, we do not use it for training. This creates a training corpus of 1,203 documents. 


\begin{figure}[H]
\begin{center}
\includegraphics[width=3.4in]{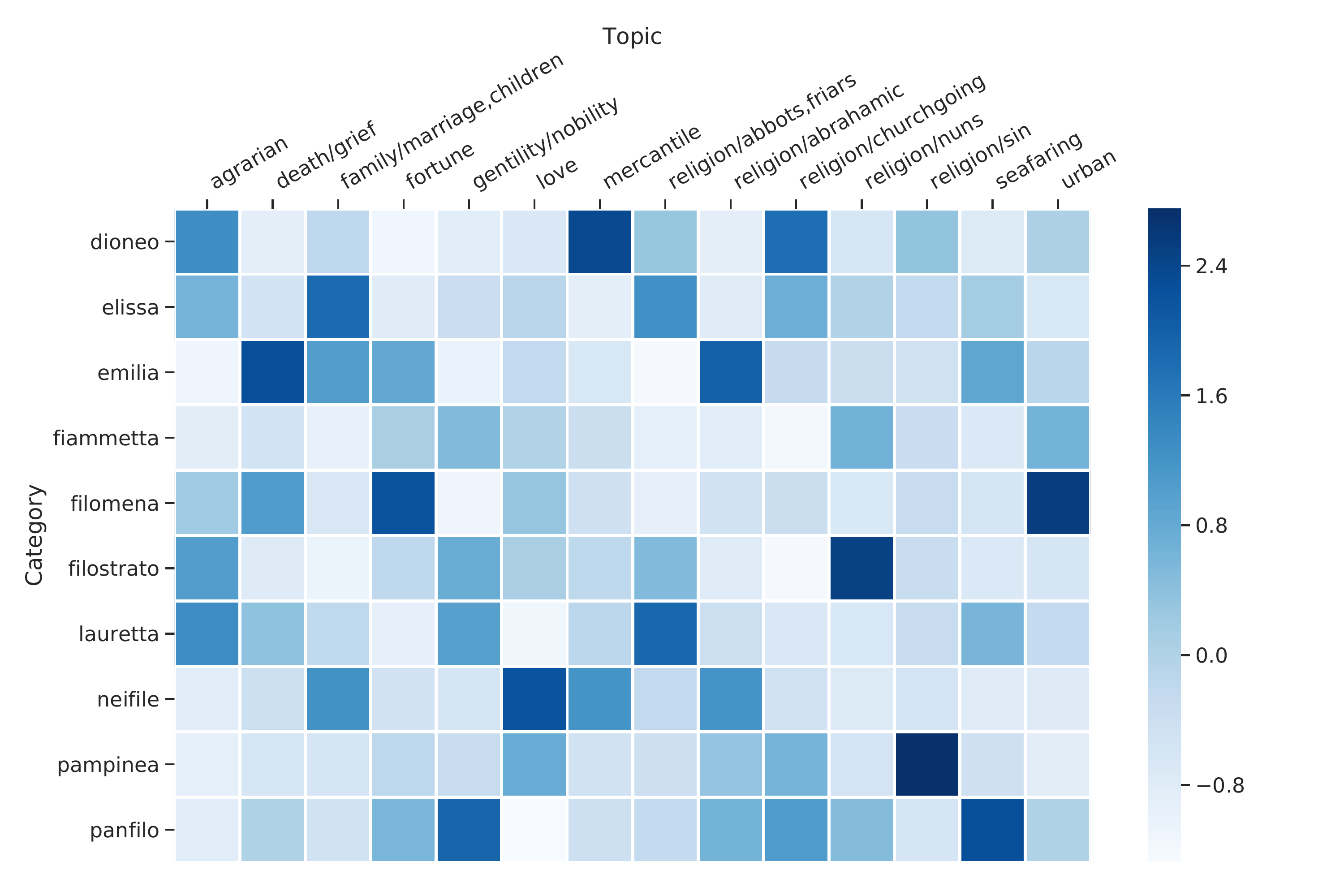}
\end{center}

\caption{Each storyteller according to their underlying distribution of topics. Columns are normalized to highlight differences across topics.}
\label{fig:narrators_heatmap}
\end{figure}

We manually validated the quality of the resulting topics to see if they were semantically meaningful, and we were able to determine some clear themes. For example, one topic's top words include \emph{nave} (ship), \emph{mare} (sea), \emph{isola} (island), \emph{barca} (boat), and \emph{vento} (wind), to which we assigned overall topic designation \texttt{seafaring}. Of the 20 topics, 14 had very clear semantic themes, while the remaining 6 were more illusive. 
Therefore, to achieve a clearer picture of the variation over storytellers, we removed these 6 topics in our plots.
We then validated the remaining topics at the \emph{novella} level, plotting a heat map in which each each row is a \emph{novella} topic distribution. This heat map enabled us to spot-check if particular \emph{novelle} had reasonable topic distributions.\footnote{This heatmap is available at \url{https://github.com/pasta41/decameron}.}  

Our storyteller-topic results are summarized in Figure \ref{fig:narrators_heatmap}, which overall indicates that there are thematic differences between the individual storytellers. If the storytellers were truly indistinguishable, it is unlikely that we would observe variation in the topic signatures. Of particular note are cells in the heatmap that show a uniquely highly-weighted presence for a topic that is relatively absent for each of the other nine narrators. To call attention to three examples, we can see this for \texttt{religion/sin} for Pampinea, \texttt{mercantile} for Dioneo, and \texttt{seafaring} for Panfilo. 

\begin{figure}[H]
\begin{center}
\includegraphics[width=3.4in]{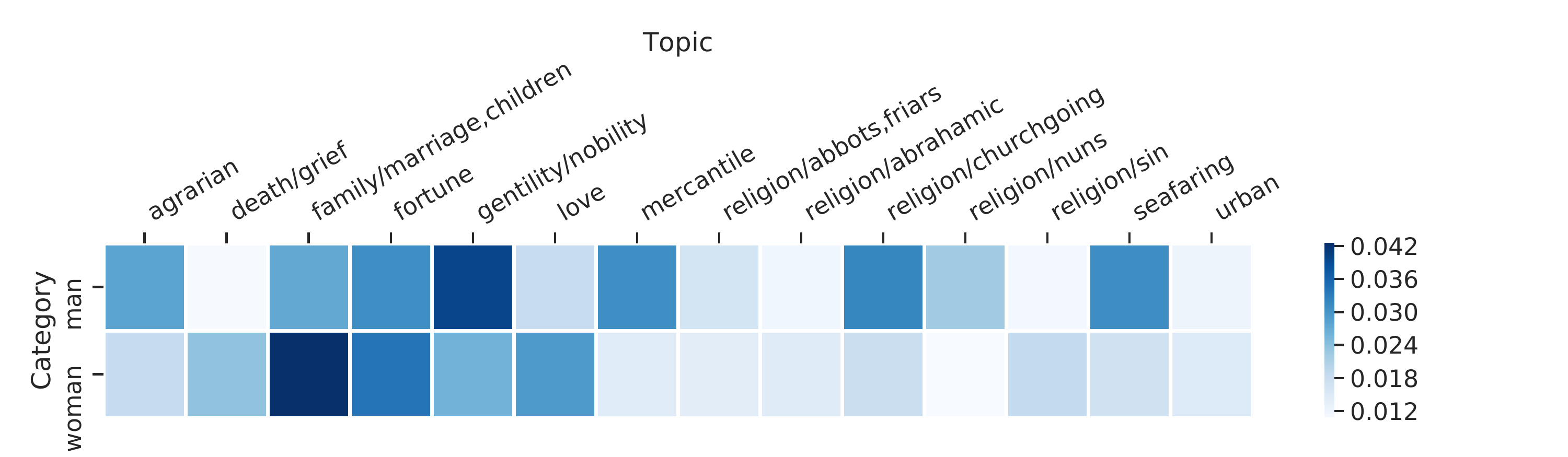}
\end{center}

\caption{Topic distributions by storyteller gender.}
\label{fig:gender_heatmap}
\end{figure}

To see another view of these results, we performed a similar analysis, in which we grouped topic distributions more coarsely---by storyteller gender instead of individual storyteller (Figure \ref{fig:gender_heatmap}). Two interesting observations for these results are that the men discuss \texttt{mercantile} themes considerably more than the women, and the women discuss \texttt{love} more than the men. Perhaps the \texttt{mercantile} results are unsurprising, given men's unrestricted ability to participate in economic endeavors---a privilege underscored in the Author's Proem~\cite{branca2014decameron}. However, the result concerning \texttt{love} is somewhat surprising. The Author's stated purpose in the Proem is to relieve the suffering of women in love, and the three men are said to be in love with three of the women of the \emph{brigata}, so it may seem unusual for words associated with love to be more strongly collocated with women than with men.

\section{Conclusion and Future Work}

While our work has focused on a specific question---whether the members of the \emph{Decameron}'s \emph{brigata} exhibit distinct storytelling personalities---we have illustrated broader lessons for small-text, specialized-language digital humanities scholarship. 
A central tension for work in low-resource domains is whether to focus on building tools and resources to mimic large, English-language resources or to instead work around the lack of resources by relying on methods that do not require much training data. While we have taken the latter path in this paper, we see ample opportunities for both developing models and annotating larger datasets for this domain~\citep{bai2021pretrain}. For example, while medieval Italian is syntactically quite different from modern Italian, some linguistic studies on specific texts indicate significant lexical overlap.\footnote{\citet{demauro2001dante} points out that if we examine the fundamental vocabulary of Italian (i.e., the most common 2000 words) we find that 92\% of them are words that Dante used in his \textit{Divina Commedia}.} 
Based on this observation, future work could modify existing Italian contextual models for high-fidelity use on medieval Italian works.

For scholars of the \textit{Decameron}, our highlighted results indicate areas for further inquiry. 
For example, a close-reading analysis of the \emph{novelle} could explain when and why the women storytellers talk about more ``male'' topics (e.g., \texttt{mercantile} themes) and would complement our topic modeling results. More broadly, our release of a simplified format of the digitized text will facilitate future digital humanities research related to Boccaccio's \emph{Decameron}. 



\bibliography{references}

\begin{thebibliography}{38}
\expandafter\ifx\csname natexlab\endcsname\relax\def\natexlab#1{#1}\fi

\bibitem[{Antoniak(2021)}]{antoniak2021mallet}
Maria Antoniak. 2021.
\newblock \href {https://github.com/maria-antoniak/little-mallet-wrapper}
  {little-mallet-wrapper}.

\bibitem[{Bai et~al.(2021)Bai, Ritter, and Xu}]{bai2021pretrain}
Fan Bai, Alan Ritter, and Wei Xu. 2021.
\newblock Pre-train or annotate? domain adaptation with a constrained budget.
\newblock In \emph{Proceedings of the 2021 Conference on Empirical Methods in
  Natural Language Processing}. Association for Computational Linguistics.

\bibitem[{Bamman et~al.(2013)Bamman, O{'}Connor, and
  Smith}]{bamman2013learning}
David Bamman, Brendan O{'}Connor, and Noah~A. Smith. 2013.
\newblock \href {https://www.aclweb.org/anthology/P13-1035} {Learning latent
  personas of film characters}.
\newblock In \emph{Proceedings of the 51st Annual Meeting of the Association
  for Computational Linguistics (Volume 1: Long Papers)}, pages 352--361,
  Sofia, Bulgaria. Association for Computational Linguistics.

\bibitem[{Blei et~al.(2003)Blei, Ng, and Jordan}]{blei2003lda}
David~M. Blei, Andrew~Y. Ng, and Michael~I. Jordan. 2003.
\newblock \href {http://jmlr.org/papers/v3/blei03a.html} {{Latent Dirichlet
  Allocation}}.
\newblock \emph{J. Mach. Learn. Res.}, 3:993--1022.

\bibitem[{Boccaccio(1995)}]{mcwilliam1975boccaccio}
Giovanni Boccaccio. 1995.
\newblock \emph{Decameron}, 2 edition.
\newblock Penguin Books, London, England.

\bibitem[{Boccaccio(2014)}]{branca2014decameron}
Giovanni Boccaccio. 2014.
\newblock \emph{Decameron, edited by Vittore Branca}.
\newblock Einaudi, Torino.

\bibitem[{Branca(1975)}]{branca1975boccaccio}
Vittore Branca. 1975.
\newblock \emph{Boccaccio medievale}, 4 edition.
\newblock G. C. Sansoni, Firenze.

\bibitem[{Branca(2003)}]{decameron2003online}
Vittore Branca. 2003.
\newblock \href
  {http://backend.bibliotecaitaliana.it/wp-json/muruca-core/v1/xml/bibit00026}
  {The {D}ecameron}.
\newblock Digitized by the Sapienza University of Rome, Biblioteca italiana
  Project.

\bibitem[{Brooke et~al.(2015)Brooke, Hammond, and Hirst}]{brooke2015wasteland}
Julian Brooke, Adam Hammond, and Graeme Hirst. 2015.
\newblock \href {https://aclanthology.org/2015.lilt-12.2} {{Distinguishing
  Voices in The Waste Land using Computational Stylistics}}.
\newblock In \emph{Linguistic Issues in Language Technology, Volume 12, 2015 -
  Literature Lifts up Computational Linguistics}. CSLI Publications.

\bibitem[{{Brown University Italian Studies Department}()}]{decameronweb}
{Brown University Italian Studies Department}.
\newblock \href {https://www.brown.edu/Departments/Italian_Studies/dweb/}
  {{Decameron Web}}.

\bibitem[{Chambers and Jurafsky(2009)}]{chambers2009unsupervised}
Nathanael Chambers and Dan Jurafsky. 2009.
\newblock \href {https://www.aclweb.org/anthology/P09-1068} {Unsupervised
  learning of narrative schemas and their participants}.
\newblock In \emph{Proceedings of the Joint Conference of the 47th Annual
  Meeting of the {ACL} and the 4th International Joint Conference on Natural
  Language Processing of the {AFNLP}}, pages 602--610, Suntec, Singapore.
  Association for Computational Linguistics.

\bibitem[{Dardano(2012)}]{dardano2012sintassi}
Maurizio Dardano, editor. 2012.
\newblock \emph{Sintassi dell'italiano antico}.
\newblock Carocci, Roma.
\newblock Vol. 1, La prosa del Duecento e del Trecento.

\bibitem[{{De Mauro}(2001)}]{demauro2001dante}
Tullio {De Mauro}. 2001.
\newblock {Dante, il gendarme e l'articolo 3 della Costituzione}.
\newblock In \emph{Dante, il gendarme e la bolletta. La communicazione pubblica
  in Italia e la nuova bolletta}, pages 3--11. Laterza, Bari.

\bibitem[{Devlin et~al.(2019)Devlin, Chang, Lee, and
  Toutanova}]{devlin2019bert}
Jacob Devlin, Ming-Wei Chang, Kenton Lee, and Kristina Toutanova. 2019.
\newblock \href {https://doi.org/10.18653/v1/N19-1423} {{BERT}: Pre-training of
  deep bidirectional transformers for language understanding}.
\newblock In \emph{Proceedings of the 2019 Conference of the North {A}merican
  Chapter of the Association for Computational Linguistics: Human Language
  Technologies, Volume 1 (Long and Short Papers)}, pages 4171--4186,
  Minneapolis, Minnesota. Association for Computational Linguistics.

\bibitem[{Findlen(2020)}]{findlen2020covid}
Paula Findlen. 2020.
\newblock \href
  {https://bostonreview.net/arts-society/paula-findlen-what-would-boccaccio-say-about-covid-19}
  {{What Would Boccaccio Say About COVID-19?}}
\newblock \emph{The Boston Review}.

\bibitem[{Geman and Geman(1984)}]{geman1984gibbs}
Stuart Geman and Donald Geman. 1984.
\newblock \href {https://doi.org/10.1109/TPAMI.1984.4767596} {{Stochastic
  Relaxation, Gibbs Distributions, and the Bayesian Restoration of Images}}.
\newblock \emph{IEEE Transactions on Pattern Analysis and Machine
  Intelligence}, PAMI-6(6):721--741.

\bibitem[{Genette(1983)}]{genette1983focalization}
G\'erard Genette. 1983.
\newblock \emph{Narrative Discourse: An Essay in Method}.
\newblock Cornell University Press, Ithaca, New York.
\newblock Translation (Jane E. Lewin) of Discours du r\'ecit, a portion of the
  3rd vol. of the author's Figures, essais.

\bibitem[{Ginsberg(2015)}]{ginsberg2015tellers}
Warren Ginsberg. 2015.
\newblock \emph{{Tellers, tales, and translation in Chaucer's Canterbury
  Tales}}.
\newblock Oxford University Press, Oxford, United Kingdom; New York, NY.

\bibitem[{Goyal et~al.(2010)Goyal, Riloff, and
  Daum{\'e}~III}]{Goyal2010AutomaticallyPP}
Amit Goyal, Ellen Riloff, and Hal Daum{\'e}~III. 2010.
\newblock \href {https://www.aclweb.org/anthology/D10-1008} {Automatically
  producing plot unit representations for narrative text}.
\newblock In \emph{Proceedings of the 2010 Conference on Empirical Methods in
  Natural Language Processing}, pages 77--86, Cambridge, MA. Association for
  Computational Linguistics.

\bibitem[{Grossi(1991)}]{grossi1991filomena}
Paolo Grossi. 1991.
\newblock {Per una rivalutazione dei narratori del Decameron: Filomena e la
  novella di Lisabetta (Decameron IV, 5)}.
\newblock \emph{Critica letteraria}, 19:145--57.

\bibitem[{Hoover(2004)}]{hoover2004testing}
David~L Hoover. 2004.
\newblock Testing {B}urrows's delta.
\newblock \emph{Literary and Linguistic Computing}, 19(4):453--475.

\bibitem[{Iyyer et~al.(2016)Iyyer, Guha, Chaturvedi, Boyd-Graber, and
  Daum{\'e}~III}]{iyyer2016feuding}
Mohit Iyyer, Anupam Guha, Snigdha Chaturvedi, Jordan Boyd-Graber, and Hal
  Daum{\'e}~III. 2016.
\newblock \href {https://doi.org/10.18653/v1/N16-1180} {Feuding families and
  former {F}riends: Unsupervised learning for dynamic fictional relationships}.
\newblock In \emph{Proceedings of the 2016 Conference of the North {A}merican
  Chapter of the Association for Computational Linguistics: Human Language
  Technologies}, pages 1534--1544, San Diego, California. Association for
  Computational Linguistics.

\bibitem[{Kittredge(1915)}]{kittredge1915chaucer}
George~Lyman Kittredge. 1915.
\newblock \emph{Chaucer and His Poetry}.
\newblock Harvard University Press, Cambridge, Massachusetts.

\bibitem[{Lawton(1985)}]{lawton1985chaucer}
David Lawton. 1985.
\newblock \emph{Chaucer's Narrators}.
\newblock D.S. Brewer, Suffolk, UK.

\bibitem[{Marafioti(2001)}]{marafioti2001lauretta}
Martin Marafioti. 2001.
\newblock \href {https://doi.org/10.1179/itc.2001.19.2.7} {{Boccaccio's
  Lauretta: The Brigata's Bearer of Bad News}}.
\newblock \emph{Italian Culture}, 19(2):7--18.

\bibitem[{Marcus(1979)}]{marcus1979form}
Millicent Marcus. 1979.
\newblock \emph{An Allegory of Form: Literary Self-consciousness in the
  Decameron}.
\newblock Anma Libri.

\bibitem[{Marcus(2020)}]{marcus2020covid}
Millicent Marcus. 2020.
\newblock \href
  {https://yalereview.yale.edu/reading-decameron-through-lens-covid-19}
  {{Reading the 'Decameron' Through the Lens of COVID-19: The Fallacy of
  Literary Distancing}}.
\newblock \emph{The Yale Review}.

\bibitem[{McCallum(2002)}]{mccallum2002MALLET}
Andrew~Kachites McCallum. 2002.
\newblock \href {http://mallet.cs.umass.edu} {{MALLET: A Machine Learning for
  Language Toolkit}}.

\bibitem[{Migiel(2004)}]{migiel2004rhetoric}
Marilyn Migiel. 2004.
\newblock \emph{A Rhetoric of the Decameron}.
\newblock The University of Toronto Press.

\bibitem[{Migiel(2015)}]{migiel2015ethical}
Marilyn Migiel. 2015.
\newblock \emph{The Ethical Dimension of the 'Decameron'}.
\newblock The University of Toronto Press.

\bibitem[{Pichotta and Mooney(2016)}]{Pichotta2016LearningSS}
Karl Pichotta and Raymond Mooney. 2016.
\newblock \href {https://ojs.aaai.org/index.php/AAAI/article/view/10347}
  {Learning statistical scripts with lstm recurrent neural networks}.
\newblock \emph{Proceedings of the AAAI Conference on Artificial Intelligence},
  30(1).

\bibitem[{Polignano et~al.(2019)Polignano, Basile, de~Gemmis, Semeraro, and
  Basile}]{polignano2019alberto}
Marco Polignano, Pierpaolo Basile, Marco de~Gemmis, Giovanni Semeraro, and
  Valerio Basile. 2019.
\newblock \href
  {https://www.scopus.com/inward/record.uri?eid=2-s2.0-85074851349&partnerID=40&md5=7abed946e06f76b3825ae5e294ffac14}
  {{AlBERTo: Italian BERT Language Understanding Model for NLP Challenging
  Tasks Based on Tweets}}.
\newblock In \emph{Proceedings of the Sixth Italian Conference on Computational
  Linguistics (CLiC-it 2019)}, volume 2481. CEUR.

\bibitem[{Prime et~al.(2020)Prime, Benoist, Orange, and
  Danticat}]{nyt2020decameron}
Kelly Prime, Mike Benoist, Tommy Orange, and Edgwidge Danticat. 2020.
\newblock \href
  {https://www.nytimes.com/2020/07/12/podcasts/the-daily/the-decameron-projec.html}
  {{The Sunday Read: 'The Decameron Project'}}.

\bibitem[{Richardson(1978)}]{richardson1978ghibelline}
Brian Richardson. 1978.
\newblock {The 'Ghibelline' Narrator in the Decameron"}.
\newblock \emph{Italian Studies}, 33:20 -- 28.

\bibitem[{Rosen-Zvi et~al.(2004)Rosen-Zvi, Griffiths, Steyvers, and
  Smyth}]{rosenzvi2004authortopic}
Michal Rosen-Zvi, Thomas Griffiths, Mark Steyvers, and Padhraic Smyth. 2004.
\newblock {The Author-Topic Model for Authors and Documents}.
\newblock In \emph{Proceedings of the 20th Conference on Uncertainty in
  Artificial Intelligence}, UAI '04, page 487–494, Arlington, Virginia, USA.
  AUAI Press.

\bibitem[{Salvi and Renzi(2010)}]{salvi2010antico}
Giampaolo Salvi and Lorenzo Renzi. 2010.
\newblock \emph{{$\: \: \:$ Grammatica} dell'italiano antico}.
\newblock Il Mulino, Bologna.

\bibitem[{{Text Encoding Initiative}(2002)}]{tei}
{Text Encoding Initiative}. 2002.
\newblock \href {https://www.tei-c.org/Vault/P4/doc/html/ref-TEI2.html} {{The
  XML Version of the TEI Guidelines: <TEI.2>}}.

\bibitem[{Wang and Iyyer(2019)}]{wang2019casting}
Shufan Wang and Mohit Iyyer. 2019.
\newblock \href {https://doi.org/10.18653/v1/N19-1130} {{C}asting {L}ight on
  {I}nvisible {C}ities: {C}omputationally {E}ngaging with {L}iterary
  {C}riticism}.
\newblock In \emph{Proceedings of the 2019 Conference of the North {A}merican
  Chapter of the Association for Computational Linguistics: Human Language
  Technologies, Volume 1 (Long and Short Papers)}, pages 1291--1297,
  Minneapolis, Minnesota. Association for Computational Linguistics.

\end{thebibliography}
\bibliographystyle{acl_natbib}

\end{document}